\documentclass[letterpaper, 10 pt, conference, final]{ieeeconf}  

\IEEEoverridecommandlockouts                              

\usepackage[LY1]{fontenc}
\usepackage[utf8]{inputenc}

\usepackage{algorithm}
\usepackage{algorithmic}

\usepackage{microtype}
\usepackage{booktabs} 
\usepackage{comment}

\usepackage{url}            
\usepackage{amsfonts}       
\usepackage{nicefrac}       

\makeatletter
\let\NAT@parse\undefined
\makeatother
\usepackage{hyperref}

\title{\LARGE \bf
Simulation-Based Reinforcement Learning \\ for Real-World Autonomous Driving
}

\author{Błażej Osiński$^{1,3}$,
Adam Jakubowski$^{1}$,
Paweł Zięcina$^{1}$,
Piotr Miłoś$^{1,5}$,\\
Christopher Galias$^{1,4}$,
Silviu Homoceanu$^{2}$ and
Henryk Michalewski$^{3}$
\thanks{$^{1}$ deepsense.ai, Warsaw, Poland}%
\thanks{$^{2}$ Volkswagen AG, Wolfsburg, Germany}%
\thanks{$^{3}$ University of Warsaw, Warsaw, Poland}%
\thanks{$^{4}$ Jagiellonian University, Cracow, Poland}%
\thanks{$^{5}$ Institute of Mathematics of the Polish Academy of Sciences, Warsaw, Poland}%
}

\usepackage{microtype}
\usepackage{graphicx}
\usepackage{booktabs} 
\usepackage{comment}
\usepackage{wrapfig}

\usepackage{hyperref}

\newcommand{\googlesitesurl}[0]{\url{http://bit.ly/34xh7z4}}
\newcommand{\bitlyurl}[0]{\url{ https://bit.ly/2k8syvh}}
\newcommand{\googlesitesurlpar}[1]{\url{https://sites.google.com/deepsense.ai/sim2real-carla/\##1}}

\usepackage[font=footnotesize]{caption}
\usepackage{graphicx}
\usepackage{booktabs}
\usepackage{float}

\newcommand{\ediscplain}{\texttt{DISCRETE-PLAIN}} 
\newcommand{\ediscreg}{\texttt{DISCRETE-REG}}

\newcommand{\econtplain}{\texttt{CONTINUOUS-PLAIN}}
\newcommand{\econtreg}{\texttt{CONTINUOUS-REG}}
\newcommand{\econtlow}{\texttt{CONTINUOUS-LOW-RAND}}
\newcommand{\econtseg}{\texttt{SEMSEG-ONLY}}
\newcommand{\ediscway}{\texttt{WAYPOINTS-DISCRETE}}
\newcommand{\econtway}{\texttt{WAYPOINTS-CONTINUOUS}}

\newcommand{\econtauxdepth}{\texttt{AUXILIARY-DEPTH}}
\newcommand{\edynamicsrnn}{\texttt{DYNAMICS-RAND-RNN}}
\newcommand{\edynamicsffw}{\texttt{DYNAMICS-RAND-FFW}}

\begin{document}

\maketitle
\thispagestyle{empty}
\pagestyle{empty}

\begin{abstract}


 
We use reinforcement learning in simulation to obtain a driving system controlling a full-size real-world vehicle. The driving policy takes RGB images from a single camera and their semantic segmentation as input. We use mostly synthetic data, with labelled real-world data appearing only in the training of the segmentation network.

Using reinforcement learning in simulation and synthetic data is motivated by lowering costs and engineering effort.

In real-world experiments we confirm that we achieved successful sim-to-real policy transfer. Based on the extensive evaluation, we analyze how design decisions about perception, control, and training impact the real-world performance.

\end{abstract}

\makeatletter
\renewcommand{\theparagraph}{%
 \ifnum\c@subsubsection<1 \thesubsection-\alph{paragraph}%
 \else \thesubsubsection.\alph{paragraph}%
 \fi%
}
\makeatother


\section{Introduction}

This work focuses on verifying whether it is possible to obtain a driving system using data from a simulator, which can be deployed on a real car.
Using synthetic data, in comparison to collecting it in the real world, reduces the cost of developing such a system.
Our policies were trained using reinforcement learning (RL) and confirmed to be useful in driving a real, full-sized passenger vehicle with state-of-the-art equipment required for Level 4 autonomy.
The real-world tests consisted of $9$ driving scenarios of total length of about $2.5$ km.



The driving policy is evaluated by its real-world performance on multiple scenarios outlined in Section~\ref{subsec:rewards}. To complete a scenario, the driving agent needs to execute from $250$ to $700$ actions at $10$ Hz at speeds varying from $15$ to $30$ km/h. In some of our experiments, the learned controller outputs the steering command directly. In other, the controller outputs waypoints which are transformed to steering commands using a proprietary control system.
In this work, we decided to limit intermediate human-designed or learned representations of the real world only to semantic segmentation. The semantic segmentator used in our system is the only component trained using the real-world data -- its training process mixes real-world and synthetic images.
Our driving policies are trained only in simulation and directly on visual inputs, understood as RGB images along with their segmentation masks. The input contains also selected car metrics and a high-level navigation command inspired by \cite{cond_imit}.


Using reinforcement learning and RGB inputs was a deliberate design decision. The goal behind this choice was to answer the following research questions: Is it possible to train a driving policy in an end-to-end fashion? Is such a policy obtained in simulation capable of real-world driving? 

Using simulation and reinforcement learning is instrumental in generating rich experience. The former enables avoiding hard safety constraints of real-world scenarios and lets the latter to explore beyond the imagination of hand-designed situations. Furthermore, RL enables end-to-end training with none or little human intervention. This property is highly desirable as it greatly reduces the engineering effort. It may also eliminate errors arising when gluing a heterogeneous system consisting of separate perception and control modules.

The quality of a simulator is crucial to obtain transferable polices. We use CARLA \cite{carla}, which offers reasonable fidelity of physical and visual layers. This fidelity comes at a high computation cost, however. In our case, it was alleviated by implementing a parallelized training architecture inspired by IMPALA \cite{impala}. With our current infrastructure, we gathered as much as $100$ years of simulated driving experience. This enabled us to conduct several lines of experiments testing various design choices. These experiments constitute the main contribution of this work:  

\noindent
{\bf 1. In simulation:}
we verify the influence of visual randomizations on transfer between different scenarios in simulation; results are summarized in Section \ref{subsec:results_sim}.

\noindent
{\bf 2. In the real world:} we test $10$ models listed in Table \ref{table:experiments} on $9$ driving scenarios. In total we report results gathered over more than $400$ test drives. See Section \ref{subsec:results_real} for a detailed description.


\begin{table}[htb]
\begin{center}
\resizebox{\columnwidth}{!}{
\begin{tabular}{|c|c|}
\hline
model & description \\ \hline
\econtplain & base experiment \\ \hline
\econtlow & experiment using smaller number of randomizations \\ \hline
\ediscplain & model using discrete actions \\ \hline
\econtreg & experiment with additional $l_2$ regularization \\ \hline
\ediscreg & analog of \ediscplain{} with additional $l_2$ regularization \\ \hline
\econtseg & model with semantic segmentation as only visual input \\ \hline
\ediscway & model driving with waypoints \\ \hline 
\econtauxdepth & model predicting depth as auxiliary task \\ \hline
\edynamicsffw & feed-forward model trained with dynamics randomizations \\ \hline
\edynamicsrnn & model with memory trained with dynamics randomizations \\ \hline
\end{tabular}}
\end{center}
\caption{Summary of models evaluated in this work.}
\label{table:experiments}
\end{table}

In Section \ref{subsec:failure} we describe two failure cases and in Section \ref{subsec:offline_models} we assess a proxy metric potentially useful for offline evaluation of models. 
We provide recordings from $9$ autonomous test drives at \bitlyurl{}. The test drives correspond to the scenarios listed in Figures~\ref{fig:routes}.


\section{Related work}
\label{sec:related_work}
\paragraph{Synthetic data and real-world robotics} Synthetic images were used in the ALVINN experiment \cite{alvinn}. \cite{cad2rl} proposed a training procedure for drones and \cite{end_to_end_imperial,assymetric,sim2real_rand, domain_rand, openai} proposed experiments with robotic manipulators where training was performed using only synthetic data. Progressive nets and data generated using simulator were used in \cite{progressive} to learn policies in the domain of real-world robot manipulation. A driving policy for a one-person vehicle was trained in \cite{wayve_no_labels}. The policy in \cite{wayve_no_labels} is reported to show good performance on a rural road and the training used mostly synthetic data generated by Unreal Engine 4. Our inclusion of segmentation as described in Section~\ref{subsec:segmentation} is inspired by sim-to-real experiments presented in \cite{sim2real_sim2sim} and \cite{modularity}. Visual steering systems inspired by \cite{cad2rl} and trained using synthetic data were presented in \cite{sim2real_invariant_sadeghi, divis}.  

\paragraph{Synthetic data and simulated robotics} Emergence of high-quality general purpose physics engines such as MuJoCo \cite{mujoco}, along with game engines such as Unreal Engine 4 and Unity, and their specialized extensions such as CARLA \cite{carla} or AirSim \cite{airsim}, allowed for creation of sophisticated photo-realistic environments which can be used for training and testing of autonomous vehicles. A deep RL framework for autonomous driving was proposed in \cite{framework_torcs} and tested using the racing car simulator TORCS.
Reinforcement learning methods led to very good performance in simulated robotics -- see, for example, solutions to complicated walking tasks in \cite{deepmind_dance,learning_to_run}. In the context of CARLA, impressive driving policies were trained using imitation learning \cite{cond_imit, deep_imitative_rowan}, affordance learning \cite{geiger}, reinforcement learning \cite{berkeley_impressive_carla}, and a combination of model-based and imitation learning methods proposed in \cite{deep_imitative_rowan}. 
However, as  stated in \cite{wayve_no_labels}: ``training and evaluating methods purely in simulation is often `doomed to succeed` at the desired task in a simulated environment'' and indeed, in our suite of experiments described in Section~\ref{sec:envs} most of the simulated tasks can be relatively easily solved, in particular when a given environment is deterministic and simulated observations are not perturbed. 

\paragraph{Reinforcement learning and real-world robotics} A survey of various applications of RL in robotics can be found in
\cite[Section 2.5]{desirenroth_survey}.
The role of simulators and RL in robotics is discussed in \cite[Section IV]{limits_and_potentials}.  
In \cite{cad2rl, assymetric,sim2real_rand, domain_rand, openai, sim2real_sim2sim, wayve_no_labels, progressive} policies are deployed on real-world robots and training is performed mostly using data generated by simulators. \cite{generalization_through_simulation} proposes a system with dynamics trained using real-world data and perception trained using synthetic data. Training of an RL policy in the TORCS engine with a real-world deployment is presented in \cite{torcs_real}.


\section{Environment and learning algorithm}
\label{sec:envs}
We use CARLA \cite{carla}, an open-source simulator for autonomous driving research based on Unreal Engine 4. CARLA features open assets, including seven built-in maps, $14$ predefined weather settings,  and multiple vehicles with different physical parameters. In our experiments we use input from simulated cameras; their settings, including position, orientation, and field of view, can be customized. Two visual quality levels (\texttt{LOW} and \texttt{EPIC}) are supported; the latter implements visual features including shadows, water reflections, sun flare effects, and antialiasing. 

Below we describe our experimental setup as used in the basic \econtplain~experiment. We varied its various elements in other experiments. Details are provided in Section~\ref{subsec:results_real}.

\begin{figure*}[h]
\vskip 0.5em
\begin{center}
\includegraphics[width=.31\linewidth]{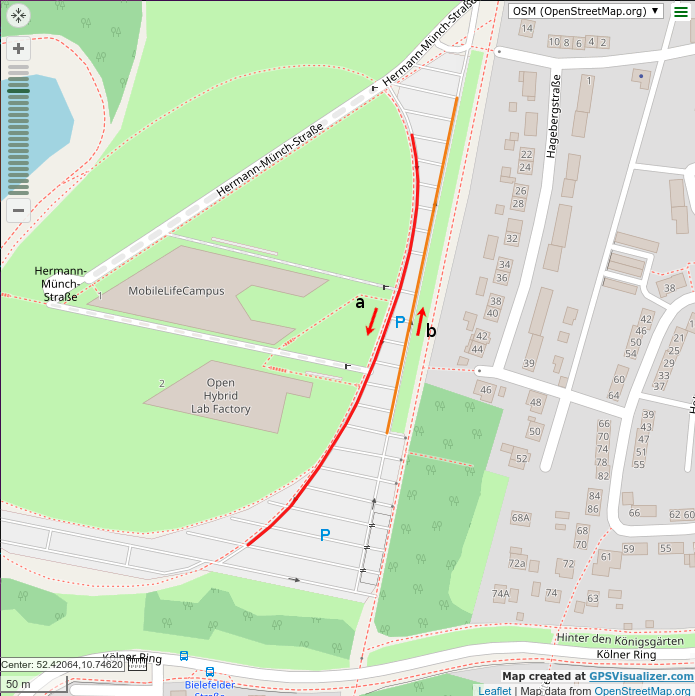}
~
\includegraphics[width=.31\linewidth]{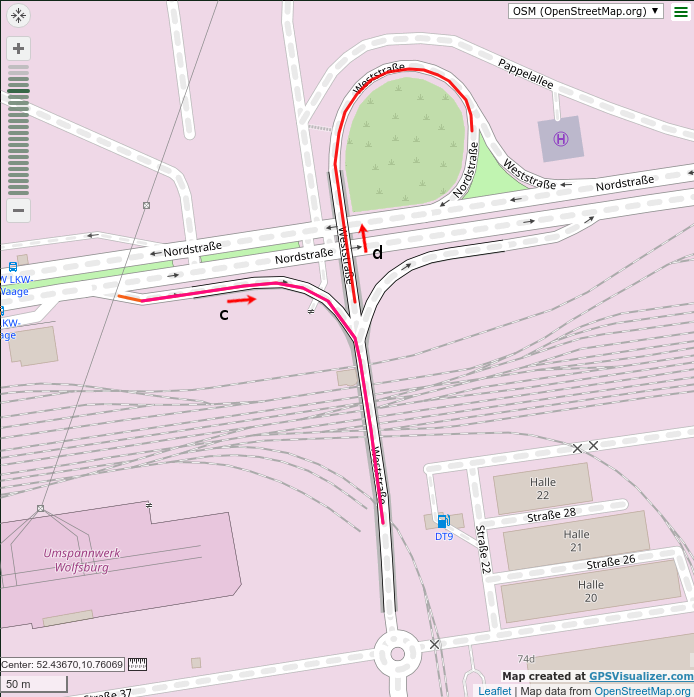}
~
\includegraphics[width=.31 \linewidth]{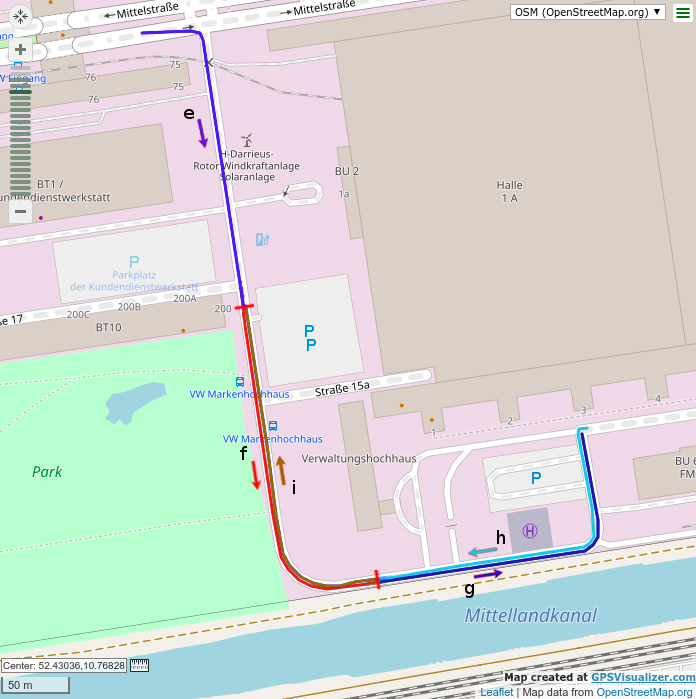}
\end{center}
 \caption{All real-world scenarios used in our experiments. Left map: (a)~\texttt{autouni-arc}, (b)~\texttt{autouni-straight}.
 Center map: (c)~\texttt{factory\_city-overpass}*, (d)~ \texttt{factory\_city-overpass\_exit}.
 Right map: (e)~\texttt{factory\_city-tunnel-bt10}*, (f)~\texttt{factory\_city-bt10-u\_turn}, (g)~\texttt{factory\_city-u\_turn-sud\_strasse},
 (h)~\texttt{factory\_city-sud\_strasse\_u\_turn}*,
 (i)~\texttt{factory\_city-u\_turn-bt10}*.
Scenarios marked with * were used for training in simulation.
}
 \label{fig:routes}
\end{figure*}

\paragraph{Simulated and real-world scenarios}
\label{subsec:rewards}
The models were tested on $9$ real-world scenarios presented in Figure~\ref{fig:routes}. These scenarios contain diverse driving situations, including turns and the entry and exit of an overpass.
For training, we developed new CARLA-compatible maps resembling the real-world testing area (approximately $50$\% of the testing scenarios were covered).
We used these maps along with maps provided in CARLA for training, with some scenarios reserved for validation.
In all scenarios the agent's goal is to follow a route from start to finish. These routes are lists of checkpoints: they are generated procedurally in CARLA maps and predefined in maps developed by us.

In training, agents are expected to drive in their own lanes, but other traffic rules are ignored.
Moreover, we assume that the simulated environment is static, without any moving cars or pedestrians, hence a number of a safety driver interventions during test deployments in real traffic is unavoidable. 
 
\begin{figure*}[h]
 \vskip 0.5em
 \centering
 \includegraphics[width=0.95\linewidth]{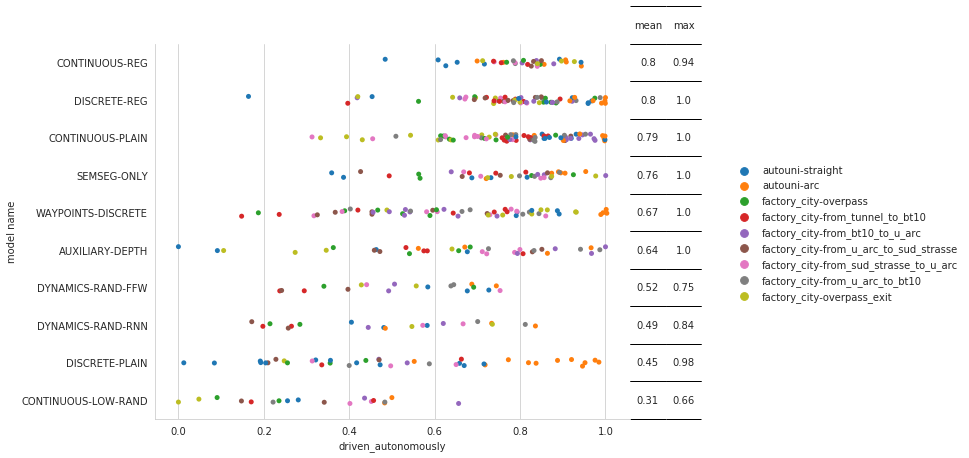}
 \caption{Summary of experiments with baselines across nine real-world scenarios. The columns to the right show the mean and max of autonomy (the percentage of distance driven autonomously). Models are sorted according to their mean performance. Print in color for better readability.}
 \label{fig:leaderboard_final}
\end{figure*}

\begin{figure*}[h]
\begin{center}     
\includegraphics[width=0.9\linewidth]{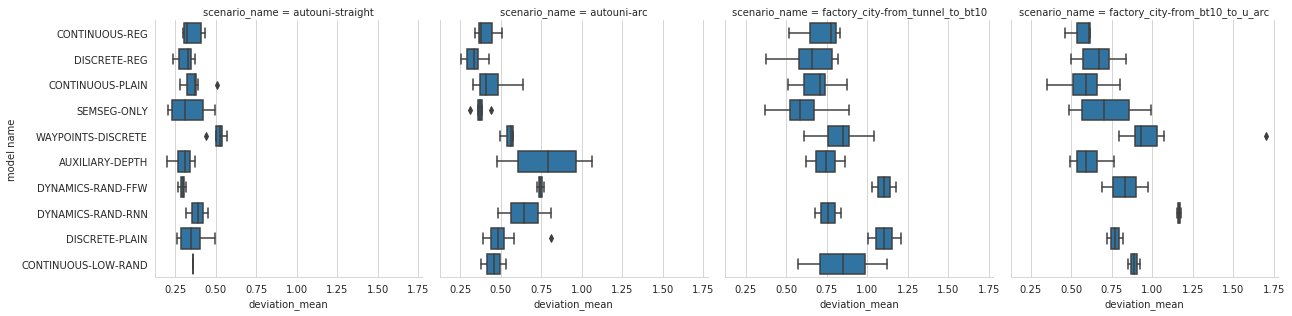}
\end{center}
 \caption{Average deviation of models from expert trajectories. Measurements based on GPS. The graphs for all scenarios can be found on the website \googlesitesurl{}}
 \label{fig:deviations}
\end{figure*}

\paragraph{Rewards in simulation and metrics of real-world performance}
In simulation, the agent is rewarded for following a reference trajectory, which provides a dense training signal. The episode fails if the agent diverges from the trajectory more than $5$ meters or collides with an obstacle. In the real world, for each scenario, we measure the percentage of distance driven autonomously (i.e. without human intervention); results are presented in Figure~\ref{fig:leaderboard_final}.
Since tests were made in an uncontrolled environment with other vehicles and pedestrians, the safety driver was instructed to take over in all situations which were potentially risky. We also measure divergence from expert trajectories (see Figure~\ref{fig:deviations}).

\paragraph{Actions}
\label{subsec:actions}
Vehicles are controlled by two values: throttle and steering. The throttle is controlled by a PID controller with speed set to a constant, and thus our neural network policies only control the steering. We explore various possibilities for action spaces. 
Unless stated otherwise, in training the policy is modeled as a Gaussian distribution over the angle of the steering wheel. In evaluations we use the mean of the distribution. 



\paragraph{Semantic segmentation}
\label{subsec:segmentation} The semantic segmentation model is trained in a supervised way separately from the reinforcement learning loop. We used the U-Net \cite{ronneberger2015unet} architecture and synthetic data from CARLA (which can render both RGB images and their ground-truth semantic labels), the Mapillary dataset \cite{MVD2017}, as well as real-world labeled data from an environment similar to the one used in test drives.
The output of the model is further simplified to include only the classes most relevant to our problem: road, road marking, and everything else (i.e. obstacles).

\paragraph{Observations}
\label{subsec:observations}
The observation provided to the agent consists of visual input and two car metrics (speed and acceleration).
In order to disambiguate certain road situations (e.g. intersections), a high-level navigation command is also given: \texttt{lane follow, turn right/left} or \texttt{go straight}. The commands resemble what could be provided by a typical GPS-based navigation system.

The visual input is based on an RGB image from a single front camera which is downscaled to $134\times 84$ pixels. The camera position and orientation in simulation was configured to reflect the real-world setup.
The RGB observation is concatenated with its semantic segmentation as described in the previous subsection. Including this component was motivated by \cite{modularity}, which claims that it
``contains sufficient information for following the road and taking turns, but it is abstract enough to support transfer''. We have further evaluated this claim in experiment \econtseg{}.


\paragraph{Domain randomizations} Randomizations are considered to be pivotal to achieve sim-to-real transfer (and robust polices in general; see e.g. \cite{openai}). In our experiments we used the following list of visual randomizations: $10$ weather settings (we used CARLA weather presets, which affect only the visual features of the environment), the simulation quality (we used both LOW and EPIC), camera input randomizations (we used a set of visual augmentations, such as adding gaussian noise, varying brightness, and applying blur or cutout \cite{cutout}). We recall that our policies are trained on multiple scenarios and different maps, which is also aimed to increase robustness. Unless stated otherwise these randomizations are used in all experiments. In a separate experiment we also evaluated using randomization of dynamics, see description in Section \ref{exp:dynamicsrand}.

\begin{figure}[h]
 \centering
 \includegraphics[width=0.9\linewidth]{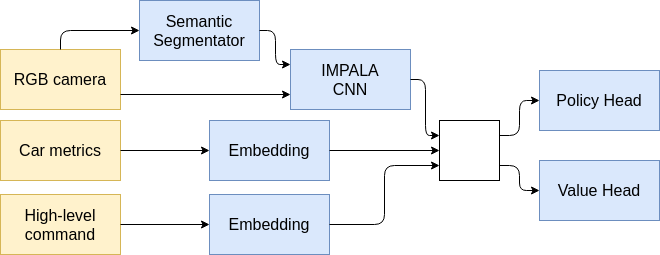}
 \caption{Network architecture.}
 \label{fig:network_arch}
\end{figure}

\paragraph{Network architecture}
The RL policy is implemented using a neural network; see its simplified architecture in Figure~\ref{fig:network_arch}.
As the feature extractor for visual input (RGB and semantic segmentation) we use the network from \cite{impala}.
Our choice was influenced by \cite{cobbe2019quantifying}, where this network was shown to generalize well to different RL environments. Note that policy transfer between simulation and reality can be seen as a generalization challenge.

\paragraph{Learning algorithm}

We used OpenAI Baselines \cite{baselines} \texttt{ppo2} (see website \googlesitesurl{} for training hyperparameters). We typically use $4$ simultaneous PPO trainers, which share gradients via Horovod \cite{horovod}. Each of them uses $2$ Tesla K40 GPUs; one for running $10$ Carla instances and the other for the optimisation of the policy network. In practice, this setup is able to gather $1.2 \cdot 10^7$ frames a day (equivalent to $13$ days of driving).

Thanks to dense rewards the training in simulation was quite stable across models and hyperparameters. 
For deployments we have decided to use $1$-$4$ models per experiment type, each trained using roughly $10^8$ frames (equivalent to about $115$ days of driving).

\paragraph{System identification}
Inspired by the importance of system identification for sim-to-real transfer demonstrated in \cite{quadruped}, we configured the CARLA simulator to mimic some values measured in the car used for deployment. These were the maximal steering angle and the time for a steering command to take effect (i.e. its delay).




\section{Experiments}
\label{sec:results}

Our models have been evaluated both in simulation and in reality, with much more focus on the latter. Below we provide detailed description of experiments conducted in both domains.

\subsection{Experiment in simulation}
\label{subsec:results_sim}

\begin{figure}[H]
\centering
\includegraphics[width=0.9\columnwidth]{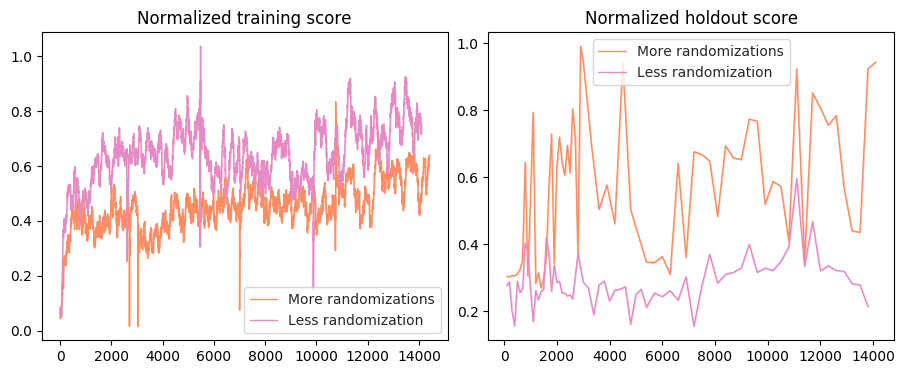}
\caption{Left: Episode scores obtained during training. The variant with less randomization is easier and faster to train. Right: On a holdout town with holdout weather better results are achieved by a model trained with more randomization.}
\label{fig:experiment_s1}
\end{figure}

In this experiment, we measure in simulation how randomizations affect performance. To this end, we apply fewer randomizations then the set used throughout all other experiments. Precisely, we used only one weather setting, trained only on the LOW quality settings and did not augment the camera inputs. We conclude that the model trained with the standard set of randomizations generalizes better to the holdout town and with the holdout weather setting (see~Figure~\ref{fig:experiment_s1}).

\subsection{Experiments in the real world}
\label{subsec:results_real}

Perhaps unsurprisingly, in our real-world experiments we observed a high level of noise. Moreover, simulation results are a poor predictor of real-world performance.  More precisely, in the set of policies above some threshold (of ``decent driving'') the correlation of simulation and real-wold scores is poor. This is perhaps the most evident instantiation of the sim-to-real gap (see also Section~\ref{subsec:offline_models} for a positive example of a offline metric correlating with real-world performance).
These factors make it hard to put forward definitive conclusions. However, we were able to observe some trends and formulate recommendations for future research. Generally, we observe the positive influence of regularization and augmentation. It also seems beneficial to have an intermediate representation layer (semantic segmentation) and branching architectures (see Section \ref{exp:waypoint}). 

Figure~\ref{fig:leaderboard_final} summarizes the performance of our models in terms of the percentage of distance \textit{driven autonomously} in all scenarios. See the project website \googlesitesurl{} for a more fine-grained presentation. Below we present a detailed per-experiment analysis.

\begin{figure}[h]
\vskip 0.5em
\begin{minipage}{0.49\linewidth}
\includegraphics[width=1.1\linewidth,trim=1.5cm 2.0cm 0.0cm 2cm,clip]{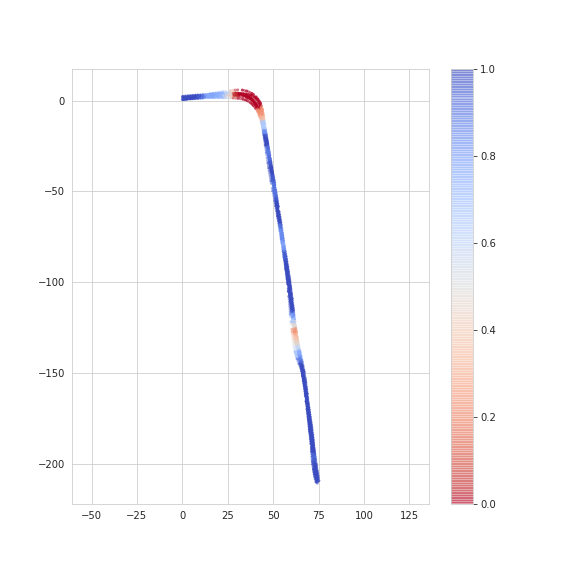}
\caption*{\econtplain{}}
\end{minipage}
\begin{minipage}{0.49\linewidth}
\includegraphics[width=1.1\linewidth,trim=1.5cm 2.0cm 0.0cm 2cm,clip]{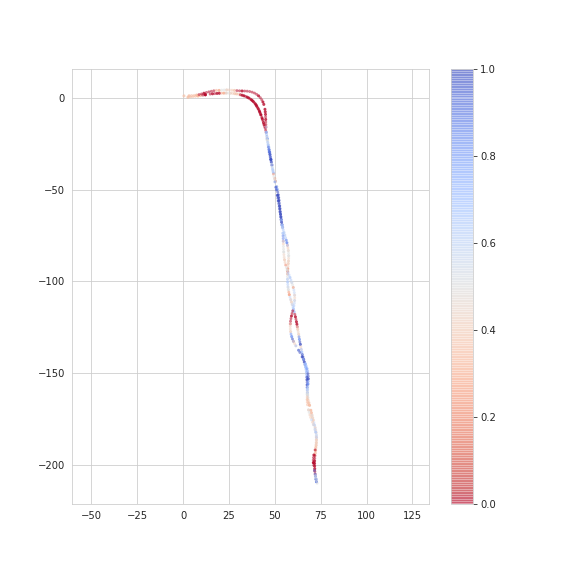}
\caption*{\ediscplain{}}
\end{minipage}
 \caption{
 Qualitative comparison of autonomy between a good model and  a lagging one. The graphs show aggregation of few trials (test drives) by these models on the route \texttt{factory\_city-tunnel-bt10}. The color depicts scale between full autonomy in all trials (blue) and human takeover in all trials (red). Graphs for other models and other routes are available on the project webpage \googlesitesurl{}.}
 \label{fig:trajectories}
\end{figure}

\paragraph{Base experiments}
The model \econtplain{} exhibited very good performance and serves as a strong baseline for comparisons with other variants discussed below. We aimed at creating a relatively simple model and training procedure. In other experiments we show that further simplifications deteriorate performance. \\
Analogously to the experiment in simulation described above, we have verified the impact of training with fewer randomizations on real-world performance. As expected, the resulting model \econtlow{} performs significantly worse, being in fact the worst model tested.

\paragraph{Discrete action space}
This experiment aimed at measuring the impact of using a discrete distribution for the action space. The steering angles were discretized into unevenly distributed atoms. More of them were placed around $0$ to improve smoothness of driving without increasing the action space (viz. [$0.$, $\pm 0.01$, $\pm 0.02$, $\pm 0.03$, $\pm  0.05$, $\pm 0.08$, $\pm 0.12$, $\pm 0.15$, $\pm 0.2$, $\pm 0.25$, $\pm 0.3$, $\pm 0.4$]; values are in radians).
During training the action was sampled, while during evaluation a deterministic policy output the expected action (i.e. the sum of the atom values multiplied by their probabilities). \\
The resulting model \ediscplain{} performed badly in real-world evaluations, mostly due to severe side-to-side wobbling. We performed more experiments with discrete actions, with results being mostly weak (see \googlesitesurl{}).

\paragraph{Regularization} Improved performance in RL generalization when using regularization was reported in \cite{cobbe2019quantifying}. We evaluated using regularization in two experiments. In the first experiment, \ediscreg{}, we fine-tuned \ediscplain{} by further training the model in a slightly altered setup: including $l_2$ regularization and reducing the policy entropy coefficient from $0.01$ to $0.001$.
The resulting model behaved significantly better (for example the wobbling observed before almost disappeared). In the second experiment, the performance of the continuous model trained with regularization -- \econtreg{} -- was only slightly improved over \econtplain{}.

\paragraph{Control via waypoints}
\label{exp:waypoint}
Following the approach presented in \cite{modularity}, in experiment \ediscway{} we trained a model to predict the next waypoint instead of steering.
Given a waypoint, low-level steering of the driving wheel is executed in order to reach this point. In simulation, this is realized by a PID controller, while in the case of the real car, we use a proprietary control system.  To ensure similar performance in simulation and reality, we limit the action space of the RL agent to waypoints reachable by both of the controllers (this consists of points within a radius of $5$ meters from the car). The action space is discrete -- potential waypoints are located every $5$ degrees between $-30$ and $30$, where $0$ is the current orientation of the vehicle. \\
In contrast to the experiments with direct steering, the continuous version of this experiment -- \econtway{} -- was weaker and exhibited strong wobbling, even in simulation.
\\
In this experiment we used a branched neural network architecture, again inspired by \cite{modularity}. Namely, we use separate heads for each of the four high-level navigation commands (see Section~\ref{subsec:observations}). Such architectures are considered to learn semantically different behaviors for the commands (e.g. \textit{go straight} vs \textit{turn left}) more easily and better handle command frequency imbalance. Our experimental results are in line with this interpretation -- our models performed turns better than ones with the standard architecture. We expect that they will offer better general performance, which we plan to investigate in further research.

\paragraph{Dynamics randomizations}
\label{exp:dynamicsrand}
In \cite{sim2real_rand, openai} dynamics randomization is pointed out as an important ingredient for successful sim-to-real transfer. 
In order to verify this in our context we introduced randomization to the following aspects of the environment: target speed, steering response (including a random multiplicative factor and bias), latency (the delay between observation and applying the policy's response to it), and noise in car metric observation (speed, acceleration, and wheel angle).
Dynamics randomization parameters were sampled once at the beginning of each training episode. \\
For both experiments with dynamics randomization -- \edynamicsrnn{} and \edynamicsffw{} -- the performance during evaluation on the real car was substandard.
Somewhat surprisingly, the feed-forward model \edynamicsffw{} performed slightly better than \edynamicsrnn{} using a GRU memory cell \cite{gru}.
This is in contrast to literature, e.g. \cite{sim2real_rand} which highlights importance of using memory cells:
intuitively, an agent with memory should infer the dynamics parameters at the beginning of the episode and utilize them for better driving. \\
We intend to further evaluate the possibility of using dynamics randomization for sim-to-real autonomous driving in future work. As a first step we will look for an explanation of the described mediocre performance. We speculate that this might be due to poor alignment of our randomizations with real-world requirements or overfitting when using high-capacity models with memory.
\paragraph{Auxiliary depth prediction} Auxiliary tasks are an established method of improving RL training (see e.g. \cite{Unsupervised2017}). Following that, in experiment \econtauxdepth{} the model also predict the depth. The depth prediction is learned in a supervised way, along with RL training. This auxiliary task slightly speeds up the training in simulation. However, in real-world evaluations it does not improve over the baseline experiments.
\paragraph{Segmentation only}
Similarly to \cite{modularity} we test the hypothesis that semantic segmentation is a useful common representation space for sim-to-real transfer. In experiment \econtseg{} the model takes in only segmentation as input and performs only slightly weaker than the baseline.


\subsection{Selected failure cases}
\label{subsec:failure}
Besides major design decisions (as described in previous sections) there are a number of small tweaks and potential pitfalls. Below we show two examples, which we find illustrative and hopefully useful for other researches and practitioners.

\subsubsection{Single-line versus double-line road markings}

In initial experiments we have used CARLA's TOWN1 and TOWN2 maps, which feature only double-line road markings. When evaluated on real-world footage, policies trained only on the above were not sensitive to single-line road markings.
This problem was fixed after introducing our custom maps, which feature single-line road markings. 


\subsubsection{Bug in reward function resulting in driving over the curb}


Our reward function includes a term which incentivizes the agent to follow scenarios routes defined by checkpoints connected with straight lines. In one of our scenarios, based on the real-world testing area, the checkpoints were too sparse, resulting in one of the connecting lines going over a curb on a bend of the road. We noticed this only after doing a real-world test, where the car also exhibited similar behavior. This illustrates the well-known tendency of RL methods to overfit to idiosyncrasies of the reward function's design. Somewhat ironically, in this case our system overcame the sim-to-real gap and transferred the unexpected behaviour precisely. This suggest that methods with stronger generalization are required -- we would hope they would generalize from the other bends the car could drive on without touching the curb.

\subsection{Offline models evaluation}

\label{subsec:offline_models}

\begin{figure}[H]
\centering
\includegraphics[width=\linewidth]{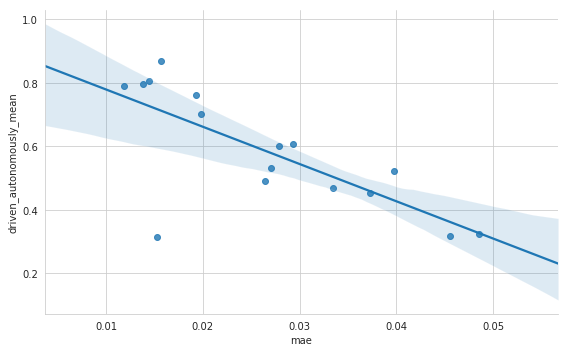}
 \caption{Dependence of the mean of the \emph{driven autonomously} metric on the mean absolute error with reference drives. Models utilizing waypoints are not included due to a different action space.}
 \label{fig:correlations2}
\end{figure}

A fundamental issue in sim-to-real experiments is that good performance in simulation does not necessarily transfer to the real-world. This is aggravated by the fact that real-world testing is costly both in time and other resources. Inspired by \cite{offlineeval} we introduced a proxy metric, which can be calculated offline and correlates with real-world evaluations. Namely, for the seven scenarios with prefix \texttt{factory\_city} we obtained a human reference drive. Frame by frame, we compared the reference steering with the one given by our models by calculating the mean absolute error. We observe a clear trend (see Figure~\ref{fig:correlations2}). While this result is still statistically rather weak, we consider it to be a promising future research direction. We present an additional offline evaluation metric ($F_1$) on the project website \googlesitesurl{}.

\section{Conclusions and future work}

We presented an overview of a series of experiments intended to train an end-to-end driving policy using the CARLA simulator. Our policies were deployed and tested on a full-size car exhibiting a substantial level of autonomy in a number of restricted driving scenarios. 

The current results let us to speculate about the following promising directions: using more regularization, control via waypoints, and using offline proxy metrics. While we obtained poor results with memory-augmented architectures, we plan to investigate the topic further.

We also consider other training algorithms which use a replay buffer such as V-trace \cite{impala} and SAC \cite{sac}. The asymmetric actor-critic architecture presented in \cite{assymetric} and a generator-discriminator pair similar to the one in \cite{graspgan} can be also beneficial for training of driving polices.  Another interesting and challenging direction is integration of an intermediate representation layer --- for example a 2D-map or a bird's-eye view, as proposed in \cite{berkeley_impressive_carla,deep_imitative_rowan,short_term_motion_pred, chauffernet}. Focusing RL training on fragments of scenarios with the highest uncertainty, see, e.g., \cite{bay_segnet} might improve driving stability. Integration of model-based methods similar to \cite{polo, simple} would be a desirable step towards better sample efficiency. 

\section{Acknowledgements}
We would like to thank Simon Barthel, Frederik Kanning, and Roman Vaclavik for their help and dedication during deployment on the physical car.
We would also like to thank Tomasz Grel and Przemysław Podczasi for their contribution to the initial stage of the project.
The work of Piotr Miłoś was supported by the Polish National Science Center grant UMO2017/26/E/ST6/00622.
The work of Henryk Michalewski was supported by the Polish National Science Center grant UMO-2018/29/B/ST6/02959.
We gratefully acknowledge Polish high-performance computing infrastructure PLGrid (HPC Centers: ACK Cyfronet AGH, PCSS) for providing computer facilities and support within computational grants no. PLG/2019/012497 and PLG/2019/012784.
We managed our experiments using \url{https://neptune.ai}. We would like to thank the Neptune team for providing us access to the team version and technical support.


\bibliography{ref}

\bibliographystyle{plain}

\end{document}